\documentclass[journal]{IEEEtran}

\usepackage{cite}

\ifCLASSINFOpdf
  \usepackage[pdftex]{graphicx}
  \graphicspath{{figures/}{../pdf/}{../jpeg/}}
  \DeclareGraphicsExtensions{.pdf,.jpeg,.png}
\else
\fi

\usepackage[utf8]{inputenc}
\usepackage[english]{babel}
\usepackage{CJKutf8}
\usepackage{amsmath}
\usepackage{array}
\usepackage{bm}
\usepackage{booktabs}
\usepackage{dblfloatfix}
\usepackage{float}
\usepackage{subcaption}
\usepackage{tipa}
\usepackage{xcolor}
\usepackage{url}

\newcolumntype{+}{>{\global\let\currentrowstyle\relax}}
\newcolumntype{^}{>{\currentrowstyle}}
\newcommand{\rowstyle}[1]{\gdef\currentrowstyle{#1}%
  #1\ignorespaces}

\begin{document}

\title{Hierarchical Character Embeddings: Learning Phonological and Semantic Representations in Languages of Logographic Origin using Recursive Neural Networks}

\author{Minh~Nguyen,
        Gia~H.~Ngo~,
        and Nancy~F.~Chen%
\thanks{
© 2019 IEEE.
Personal use of this material is permitted.
Permission from IEEE must be obtained for all other uses, in any current or future media, including reprinting/republishing this material for advertising or promotional purposes, creating new collective works, for resale or redistribution to servers or lists, or reuse of any copyrighted component of this work in other works.\protect\\
Minh Nguyen is currently with University of California - Davis, but part of the work was done at the Institute for Infocomm Research, A\textsuperscript{*}STAR.\protect\\
Gia H. Ngo is currently with Cornell University, but part of this work was done at the Institute for Infocomm Research, A\textsuperscript{*}STAR.\protect\\
Nancy F. Chen is currently with the Institute for Infocomm Research, A\textsuperscript{*}STAR.\protect\\
Nancy F. Chen is the corresponding author (nancychen@alum.mit.edu).}%
}

\maketitle

\begin{abstract}
Logographs (Chinese characters) have recursive structures (i.e.\ hierarchies of sub-units in logographs) that contain phonological and semantic information, as developmental psychology literature suggests that native speakers leverage on the structures to learn how to read.
Exploiting these structures could potentially lead to better embeddings that can benefit many downstream tasks.
We propose building hierarchical logograph (character) embeddings from logograph recursive structures using treeLSTM, a recursive neural network.
Using recursive neural network imposes a prior on the mapping from logographs to embeddings since the network must read in the sub-units in logographs according to the order specified by the recursive structures.
Based on human behavior in language learning and reading, we hypothesize that modeling logographs' structures using recursive neural network should be beneficial.
To verify this claim, we consider two tasks (1) predicting logographs' Cantonese pronunciation from logographic structures and (2) language modeling.
Empirical results show that the proposed hierarchical embeddings outperform baseline approaches.
Diagnostic analysis suggests that hierarchical embeddings constructed using treeLSTM is less sensitive to distractors, thus is more robust, especially on complex logographs.
\end{abstract}


\IEEEpeerreviewmaketitle
\section{Introduction}
\label{sec:intro}

\begin{figure}[ht]
\centering
\includegraphics[width=\linewidth]{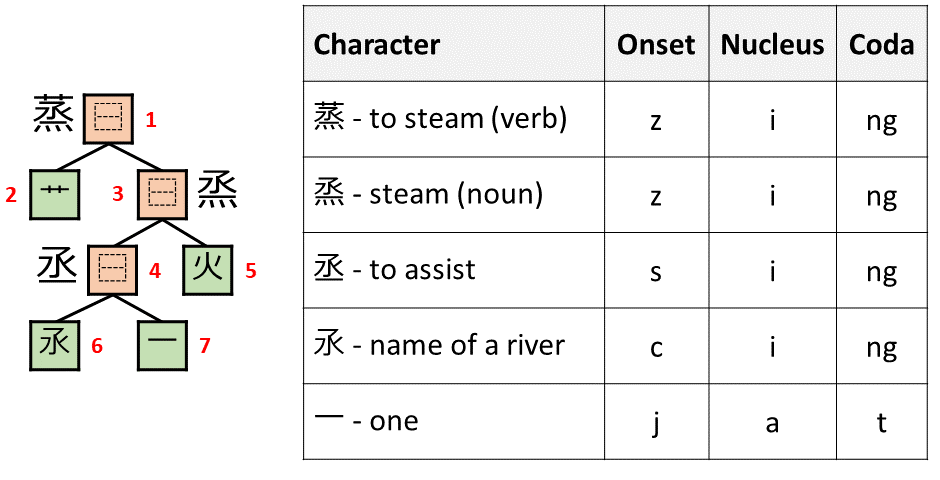}
\caption{\label{fig:example}
\textit{An example of logographic structure.
The left panel shows a binary tree representing the logograph \protect\begin{CJK*}{UTF8}{bsmi}蒸\protect\end{CJK*}.
The leaf nodes (position 2, 5, 6, 7) are sub-units forming the logograph (analogous to letters forming English words).
The inner nodes (position 1, 3, 4) are composition operators (such as vertical stacking) applied to children nodes.
The logograph \protect\begin{CJK*}{UTF8}{bsmi}蒸\protect\end{CJK*} is formed by composing all the nodes in the tree in a bottom-up fashion.
The sub-trees rooted at positions 3, 4, 5, 6, 7 also form logographs (\protect\begin{CJK*}{UTF8}{bsmi}烝, 丞, 氶, 火, 一\protect\end{CJK*}).
The right table shows the logographs' meanings and their pronunciation in Cantonese.
}}
\end{figure}

\IEEEPARstart{L}ogographic structures contain phonological and semantic information about the logographs~\cite{hsiao2006analysis}.
Language learners usually exploit logographic structures to learn logographs' pronunciation by focusing on salient sub-units of logographs that hint at pronunciations~\cite{ho1997phonological}.
Being able to focus on sub-units of logographs might explain how humans can remember the pronunciation and meanings of thousands of distinct characters.
Figure~\ref{fig:example} shows how logographic structures encode phonological and semantic information.
The \begin{CJK*}{UTF8}{bsmi}氶\end{CJK*} sub-unit (position 6) hints at the nucleus and coda in the logographs' pronunciation.
In addition, the \begin{CJK*}{UTF8}{bsmi}火\end{CJK*} sub-unit\footnote{\begin{CJK*}{UTF8}{bsmi}火\end{CJK*} is written as \begin{CJK*}{UTF8}{gkai}灬\end{CJK*} when is it at the bottom position.}
(position 5) suggests that the logographs containing this sub-unit must be related to fire.
For the four logographs \begin{CJK*}{UTF8}{bsmi}蒸, 烝, 丞, 氶\end{CJK*}, the structure of one logograph is nested within that of the preceding logograph.
For example, \begin{CJK*}{UTF8}{bsmi}烝\end{CJK*} is nested within \begin{CJK*}{UTF8}{bsmi}蒸\end{CJK*}.
Modeling this hierarchy should allow models to pick out \begin{CJK*}{UTF8}{bsmi}氶\end{CJK*} as the most relevant sub-unit for determining the pronunciation of \begin{CJK*}{UTF8}{bsmi}蒸, 烝, 丞, 氶\end{CJK*} and \begin{CJK*}{UTF8}{bsmi}火\end{CJK*} as the most relevant sub-unit for determining the semantic of \begin{CJK*}{UTF8}{bsmi}蒸\end{CJK*} and \begin{CJK*}{UTF8}{bsmi}烝\end{CJK*}.

Given the link between logographic structures and their phonology and semantics, we investigate methods to construct logograph (character) embeddings that are useful for different downstream tasks.
We consider two tasks (1) predicting logographs' Cantonese pronunciation from logographic structures and (2) language modeling.
Pronunciation prediction task requires the embeddings to contain phonological information while language modeling requires the embeddings to contain semantic information.
We propose constructing hierarchical logograph (character) embeddings of logographs from their recursive structures using treeLSTM~\cite{tai2015improved,zhu2015long}.
treeLSTM model exploits structures explicitly since it must read in sub-units in the logographs according to the order specified by the recursive structures.
We compare hierarchical embeddings against two different approaches that are commonly used to construct embeddings.
The first approach is standard embeddings~\cite{mikolov2013distributed} in which logographs are mapped to representations without utilizing the logographs' structures.
The second approach is to construct logograph embeddings from linearized structures using LSTM~\cite{graves2013generating}.
The second approach only exploits structures implicitly since the structural information is in the input data and not in the model.
Without a lot of training data, this approach is prone to overfitting and learning solutions that may not generalize well.

Modeling structures is expected to help models generalize better especially when there is limited training data~\cite{ngo2014minimal,ngo2015phonology,ngo2019phonology}.
Modeling structures has led to improvement in multiple tasks such as machine translation~\cite{yamada2001syntax,eriguchi2016tree}, sentiment analysis~\cite{tai2015improved,miyazaki2017japanese}, natural language inference~\cite{bowman2016fast}, and parsing~\cite{dyer2016recurrent,zhang2016top}.
Despite these successes, there are also cases whereby there is little improvement~\cite{li2015tree,lan2018toolkit}.
The lack of improvement could be due to either (1) the models cannot exploit structures effectively or (2) the structures do not provide information relevant to the tasks.
Thus, it is important to ensure both the high quality of structure annotations and ability of models to exploit structures effectively so as to improve overall task performance.
However, ensuring consistently high quality annotation is not simple, especially for complex tasks where multiple annotations are plausible.
The quality of structure annotation may vary between training sets and test sets or even within examples in the training sets.
Variation in annotations of training samples may happen due to disagreement between human annotators.
Variation between annotations between training and test samples may happen when models are trained on annotations provided by humans but are tested on annotations provided by parsers that were trained to mimic human annotators.
In contrast, for logographic structures, annotations are consistent since they are constructed automatically using a rule-based parser.
The rules~\cite{morioka2008chise} are defined by human experts from the Ideographic Rapporteur Group, a committee advising the Unicode Consortium about logographs therefore the annotation should be of reasonably high quality\footnote{The Kyoto University's CHaracter Information Service Environment (CHISE) project: \url{http://www.chise.org/}}.
Hence, compared to other tasks which utilize structures, tasks involving logographs could benefit more from effective modeling of structures.

In Section~\ref{sec:model}, we introduce the model to construct the hierarchical embeddings.
We apply the proposed hierarchical character embeddings to two distinct tasks:
(1) pronunciation prediction (Section~\ref{sec:exp_ph}) focusing on a case study to isolate the effects of modeling recursive structures, and
(2) language modeling (Section~\ref{sec:exp_lm}) which is an useful auxiliary task, as it characterizes many aspects of language beyond semantics (including syntactic structure and discourse processing), and language modeling can be used to pretrain many other tasks~\cite{ramachandran2017unsupervised,peters2018deep,radford2018improving,howard2018universal,devlin2019bert},
thus, it has a lot of down-stream applications.
However, due to the multifaceted nature of the language modeling task, it is hard to analyze the result qualitatively.
Section~\ref{sec:related} discussed our work in relation to other work.

\section{Model}
\label{sec:model}

\subsection{Rule-based Parser}
\label{ssec:parser}

\begin{figure}[ht]
\centering
\includegraphics[width=\linewidth]{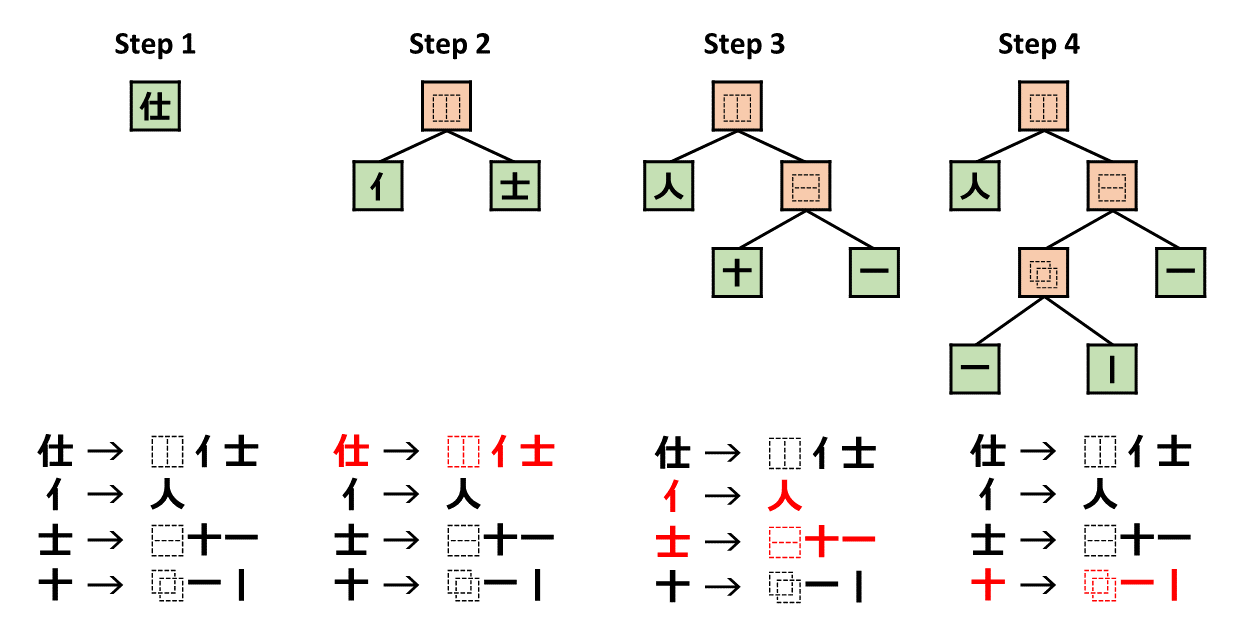}
\caption{\label{fig:parser}
\textit{Construction of logographic recursive structure using the ruled-based parser.
In this example, there are only four rules used which are shown at the bottom.
The rule used at each decomposition step is in red.
}}
\end{figure}

Decomposition of logographs into sub-units is necessary to locate the sub-units hinting at the phonetic or semantic information.
Logographs are decomposed recursively using a rule-based parser.
The substitution rules used by the parser are defined by human experts from the Ideographic Rapporteur Group.
A substitution rule defines a mapping from one logograph to sub-units and a geometric operator (Ideographic Description Character) which denotes the relative position of the sub-units.
The output of the parser is a binary tree as shown in Figure~\ref{fig:parser}.

At the start, there is only the root node, which is the logograph itself.
The parser extends the tree by recursively replacing nodes in the tree with sub-trees defined by the substitution rules.
The root of the sub-trees are the geometric operator and the children of the sub-trees are the sub-units.
This process is repeated until there is no more node in the tree that can be further decomposed.

Figure~\ref{fig:parser} shows how the structure (represented as a binary tree) is constructed for the logograph \begin{CJK*}{UTF8}{bsmi}仕\end{CJK*}.
At step 1, there is a root node \begin{CJK*}{UTF8}{bsmi}仕\end{CJK*} with no children.
At step 2, using the rule in red, the node \begin{CJK*}{UTF8}{bsmi}仕\end{CJK*} is replaced by a geometric operator (horizontal stacking) and two children nodes.
At step 3, using the rules in red, the nodes are further simplified into by \begin{CJK*}{UTF8}{bsmi}人\end{CJK*}, \begin{CJK*}{UTF8}{bsmi}十\end{CJK*} and \begin{CJK*}{UTF8}{bsmi}一\end{CJK*}.
The process terminates at step 4 where there are four leaf nodes with three distinct values \begin{CJK*}{UTF8}{bsmi}人\end{CJK*}, \begin{CJK*}{UTF8}{gkai}丨\end{CJK*}, and \begin{CJK*}{UTF8}{bsmi}一\end{CJK*} which cannot be simplified further.
There are 505 sub-units which can be leaf nodes.
These sub-units are not hand-picked, thus whether or not the representation of the sub-units carries phonetic or semantic information is automatically learned during training.
Hence, the hierarchical embeddings can be used in different tasks.
The phonetic and semantic sub-units can be at depth 1 (children of the root node) or they can reside deeper in the trees.

To construct logograph embeddings from trees, one can use bag-of-words models, sequence models, or tree-structured models~\cite{tai2015improved,zhu2015long}.
Since the ordering of sub-units within logographs is important in determining the logographs' pronunciation and meaning, order-agnostic models such as bag-of-words models are sub-optimal for constructing logograph embeddings.
Since sequence models and tree-structured models are sensitive to the ordering of sub-units, they can be used to construct logograph embeddings.
Sequence models such as recurrent neural networks (RNNs), in particular LSTM~\cite{graves2013generating}, can be used with tree inputs by first linearizing the trees into sequences.
In contrast, recursive neural networks, such as treeLSTM, can consume tree inputs directly to yield the logograph embeddings.
We compared LSTM and bi-directional LSTM (biLSTM), which are structure-agnostic, and treeLSTM, which is innate for modeling tree structures.

\subsection{Constructing Embeddings Using LSTM}
\label{ssec:lstm}

At each position $t$ in a linearized tree of length $T$, ${\bm x}_{t}$, ${\bm c}_{t}$, ${\bm h}_{t}$ are the input, cell value, and hidden state of the LSTM respectively.
The last hidden state, ${\bm h}_{T}$, is used as the logograph embedding.
Figure~\ref{subfig:lstm_model} shows the LSTM model.

\subsection{Constructing Embeddings Using Bi-directional LSTM}
\label{ssec:bilstm}

The biLSTM consists of two LSTMs, the forward LSTM and the backward LSTM, which read the linearized trees in opposite direction.
The logograph embedding, ${\bm h}_{T}$, is formed by concatenating the last hidden states of the backward and forward LSTMs, i.e. ${\bm h}^{b}_{T}$ and ${\bm h}^{f}_{T}$.

\subsection{Constructing Embeddings Using CNN}
\label{ssec:bilstm}

The model structure is similar to that of~\cite{li2018subword}.
The input to the model are also sequences formed by linearizing the trees of logographs.
The CNN model consists of 7 parallel 1D convolutional layers with kernel size from 1 to 7.
Each convolutional layer has 200 filters.
The convolutional layers are followed by max-pooling layers.
After that, the outputs are concatenated and fed through a fully-connected layer.
The output of the fully-connected layer is the logograph embedding.

\subsection{Constructing Hierarchical Embeddings Using treeLSTM}
\label{ssec:treelstm}

\begin{figure}
\centering
\begin{subfigure}{.27\textwidth}
\centering
\includegraphics[width=.9\linewidth]{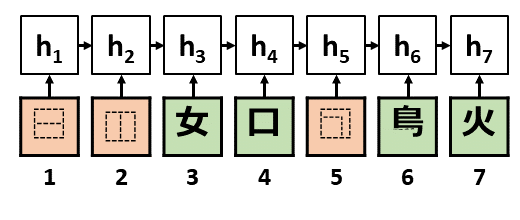}
\caption{LSTM}
\label{subfig:lstm_model}
\end{subfigure}%
\begin{subfigure}{.23\textwidth}
\centering
\includegraphics[width=.9\linewidth]{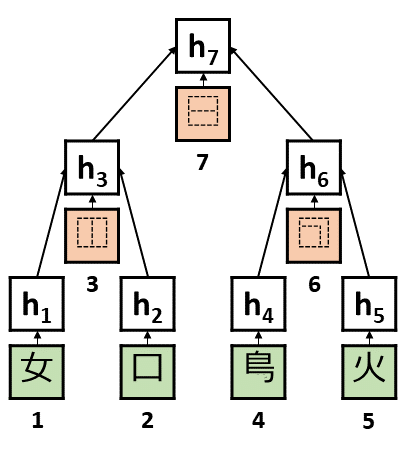}
\caption{treeLSTM}
\label{subfig:treelstm_model}
\end{subfigure}
\caption{\label{fig:models}
\textit{LSTM and treeLSTM models.
The \textbf{last} hidden layer $\bf{h}_7$ of LSTM is the logograph embedding.
The \textbf{root} hidden layer $\bf{h}_7$ of treeLSTM is the logograph hierarchical embedding.
}}
\end{figure}

At each node $n$ of the binary tree with two children $l$ and $r$, ${\bm x}_{n}$, ${\bm c}_{n}$, ${\bm h}_{n}$ are the input, cell value, and hidden state of the treeLSTM respectively.
${\bm i}$, ${\bm f}_{l}$, ${\bm f}_{r}$, ${\bm o}$ are the input gate, left forget gate, right forget gate, and output gate respectively.
The forward pass of a treeLSTM unit is given by:

\begin{eqnarray}
{\bm i} & = &
    \sigma ({\bm U}_{l}^{i} {\bm h}_{l} + {\bm U}_{r}^{i} {\bm h}_{r} +
    {\bm V}^{i} {\bm x}_{n} + {\bm V}_{l}^{i} {\bm x}_{l} + {\bm V}_{r}^{i} {\bm x}_{r})\nonumber\\
{\bm f}_{l} & = &
    \sigma ({\bm U}_{l}^{f_{l}} {\bm h}_{l} + {\bm U}_{r}^{f_{l}} {\bm h}_{r} +
    {\bm V}^{f_{l}} {\bm x}_{n} + {\bm V}_{l}^{f_{l}} {\bm x}_{l} + {\bm V}_{r}^{f_{l}} {\bm x}_{r})\nonumber\\
{\bm f}_{r} & = &
    \sigma ({\bm U}_{l}^{f_{r}} {\bm h}_{l} + {\bm U}_{r}^{f_{r}} {\bm h}_{r} +
    {\bm V}^{f_{r}} {\bm x}_{n} + {\bm V}_{l}^{f_{r}} {\bm x}_{l} + {\bm V}_{r}^{f_{r}} {\bm x}_{r})\nonumber\\
{\bm o} & = &
    \sigma ({\bm U}_{l}^{o} {\bm h}_{l} + {\bm U}_{r}^{o} {\bm h}_{r} +
    {\bm V}^{o} {\bm x}_{n} + {\bm V}_{l}^{o} {\bm x}_{l} + {\bm V}_{r}^{o} {\bm x}_{r})\nonumber\\
\tilde{{\bm c}} & = &
    \tanh ({\bm U}_{l}^{\tilde{c}} {\bm h}_{l} + {\bm U}_{r}^{\tilde{c}} {\bm h}_{r} +
    {\bm V}^{\tilde{c}} {\bm x}_{n} + {\bm V}_{l}^{\tilde{c}} {\bm x}_{l} + {\bm V}_{r}^{\tilde{c}} {\bm x}_{r})\nonumber\\
{\bm c}_{n} & = &
	{\bm i} \odot \tilde{{\bm c}} + {\bm f}_{l} \odot {\bm c}_{l} + {\bm f}_{r} \odot {\bm c}_{r}\nonumber\\
{\bm h}_{n} & = &
    {\bm o} \odot \tanh ({\bm c}_{n})\label{eq:treeLSTM}
\end{eqnarray}
The hidden state of the root node, ${\bm h}_{root}$, is considered as the representation of the entire tree.
Figure~\ref{subfig:treelstm_model} shows the treeLSTM model.

\subsection{Implementation Details}

One problem with tree-structured models is that training is very slow~\cite{irsoy2014deep}.
It is hard to batch the training samples as they might have different tree shapes~\cite{eriguchi2016tree}.
As a result, training with batch size of one is very common for tree-structured model~\cite{neubig2017fly},
which fails to maximize parallel computation and thus leads to slow training.
Instead, we used dynamic batching to speed up training and inference.
Dynamic batching in Pytorch\footnote{\url{https://devblogs.nvidia.com/recursive-neural-networks-pytorch/}} has been used to create batches of nodes on the fly to speed up the SPINN model~\cite{bowman2016fast} training and inference.
In our experiments, using a batch size of 128 results in more than 10 times faster training and inference.
Besides, we only considered binary trees and converted any ternary nodes (nodes with three children) to two nested binary nodes.
This is done to reduce the number of parameters that the model has to learn therefore improving the learning efficiency.
The tree representation is sensitive to the order of the children nodes as swapping the left child and the right child in a tree results in a character with potentially different meaning and pronunciation.
Thus, we need separate weight matrices for each of the children.
As such, modeling both binary and ternary nodes would require from 3 to 5 weight matrices whereas modeling binary trees only requires 2 weight matrices.
Since the amount of data is limited, we preferred models with fewer parameters and thus we converted all ternary nodes to binary nodes.

\section{Experiments --- Pronunciation Prediction}
\label{sec:exp_ph}
We compared embeddings produced using treeLSTM against LSTM and biLSTM.
treeLSTM operates directly on the tree form of the logograph in order to exploit the recursive structure of logographs most effectively.
In contrast, LSTM and biLSTM use more implicit structural information of the logograph in the form of linearized trees.
Since standard embeddings do not consider logographic structures, every input logograph is distinct so this approach cannot learn similarities between logographs.
Hence, we did not compare the hierarchical embeddings against standard embeddings.

\subsection{Data}

The data was extracted from UniHan database\footnote{\url{https://www.unicode.org/charts/unihan.html}}, which is a pronunciation database of characters of Han logographic languages.
Each entry consists of a character and its pronunciations in various languages such as Cantonese and Mandarin.
For entry with multiple pronunciations, since the dominant pronunciation is not indicated, we randomly picked one of the variants.
For this task, the input is the logographic character and output is the Cantonese pronunciation.
The pronunciation includes onset, nucleus, and coda.
As far as we know, lexical tones are not directly determined by logographic structures so we did not include lexical tones as prediction targets.

There are two types of logographs used in Cantonese, namely traditional and simplified characters.
Simplified characters, as the name implies, are derived from their traditional counterparts by removing or replacing some complex sub-units with simpler ones.
Non-simplified characters include both traditional characters and the subset of Chinese characters that are identical for traditional and simplified counterparts.
Hence, simplified and traditional Chinese characters are quite different in terms of unique sub-units and their complexity.

\subsection{Setup}

A common weakness of deep learning models is that they often merely memorize patterns and do not generalize well on unseen data~\cite{jia2017adversarial}.
LSTM has the same weakness as it performs well when there is abundant training data and test distribution is the same as the training distribution~\cite{lake2018generalization}.
When the test and training distributions are different, LSTM does not perform as well.
Strong generalization requires models to extrapolate to out-of-distribution data points rather than to interpolate using data points within distribution~\cite{mitchell2018extrapolation}.

To test the generalizability of standard LSTM and treeLSTM, the original UniHan dataset was split into training and test sets in three different scenarios described in Table~\ref{tbl:ph_data}.
In the first scenario, the training and test set's distribution were homogeneous: both contained traditional and simplified characters.
In the second scenario, the test set only contained simplified characters and the training set contained non-simplified characters.
In the third scenario, the distributions were different and the training data was limited: the test set contained only simplified characters while the training sets contained corresponding traditional characters.

\begin{table}[ht]
\normalsize
\centering
\begin{tabular}{+l^r^r^r}
    Scenario & Training & Validation & Test \\
\midrule
    1. Tr, Sp$ \rightarrow$ Tr, Sp & 16000 & 2400 & 2400 \\
    2. Non-Sp $\rightarrow$ Sp     & 16000 & 2400 & 2400 \\
    3. Tr $\rightarrow$ Sp         &  2302 &  200 & 2400 \\
\end{tabular}
\caption{\label{tbl:ph_data}
\textit{Number of characters (logographs) used for training and testing in each of the scenario. Tr: Traditional, Sp: Simplified}}
\end{table}

The third scenario is inspired by the fact that humans being able to predict pronunciations of simplified characters given the corresponding traditional characters, although they may rely on word context.
Given that human performance is high, it should not be impossible for models to generalize to simplified characters even when trained solely on traditional characters.
By contrasting results obtained from scenario 1 and 2, we could determine whether the models merely memorized patterns or they learned the underlying rules to predict pronunciation as humans,
since models that merely memorize patterns would do well in scenario 1 but not scenario 2.
In addition, contrasting scenario 2 and 3 would hint at how models perform in low-resource scenarios of limited training data as well as whether the bias induced by the logographic structures is useful for improving model generalization.
It should be noted that scenario 1 is the ideal case in which one is very careful in collecting data and performs data normalization.
If data is collected indiscriminately, one can end up in scenario 2.

\subsection{Task-specific Layer}
\label{ssec:task_layer}

The task-specific layer uses the logograph embedding to predict the logograph's pronunciation, which includes onset, nucleus and coda.
Probability of each sub-syllabic unit's pronunciation is given by:
\begin{eqnarray}
    CD & = & \mbox{softmax} (W^{CD} {\bm h}),\nonumber\\
    NU & = & \mbox{softmax} (W^{NU} [{\bm h}, CD]),\nonumber\\
    ON & = & \mbox{softmax} (W^{ON} [{\bm h}, CD, NU])\nonumber
\end{eqnarray}
where $W^{CD}$, $W^{NU}$ and $W^{ON}$ are weights of the fully-connected layer specific to each sub-syllabic unit.

\begin{figure}[ht]
    \centering
    \includegraphics[width=\linewidth]{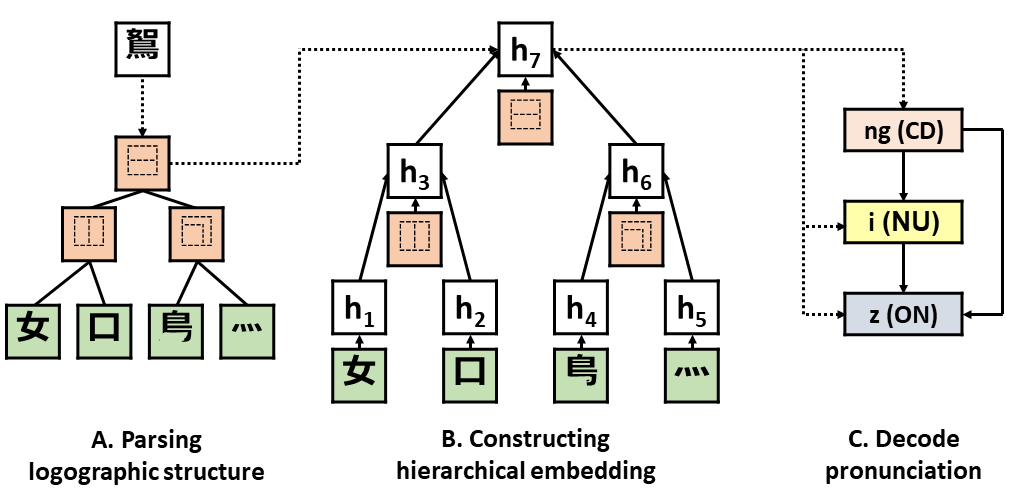}
    \caption{\label{fig:ph_model}
    \textit{Phonological prediction model using hierarchical embeddings.
    (A) The input logograph is decomposed into the logographic structure using the rule-based parser.
    (B) treeLSTM constructs hierarchical embedding from the structure.
    (C) The embedding is then used to predict the pronunciation.}
    }
\end{figure}

The setup for treeLSTM to predict a logograph's pronunciation using hierarchical embeddings is shown in Figure~\ref{fig:ph_model}.

\subsection{Metrics}

We evaluated models' performance using string error rate (SER) and token error rate (TER).
A wrongly predicted phoneme (onset, nucleus or coda) was counted as one token error.
An output containing at least one token error was counted as one string error.
We used modified Obuchowski statistical test~\cite{yang2010note} to assess the difference in predictive differences.

\subsection{Hyperparameters}

The size of hidden layers is fixed as 256.
We used dropout~\cite{srivastava14dropout} on input and hidden layers to prevent overfitting.
We optimized the models using the Adam~\cite{kingma2015adam} optimizer.
The batch size was 128.
For each of the model, we searched for the best learning rates and dropout rates using grid-search.
The learning rate ranges from $3\times10^{-2}$ to $1\times10^{-4}$.
The drop out rate ranges from 0.0 to 0.5. 

\subsection{Linearization Order}
\label{ssection:linearize}
Since there are multiple ways to linearize trees into sequences, in this section, we investigated what is the optimal linearization order for the models.
We compared three different schemes namely: in-order, pre-order, post-order linearization.
We paired each of the models (there are five models in total) with the 3 different linearization schemes.
This resulted in fifteen different combinations.
For each combination, we conducted hyperparameter search on the development set.
The lowest TER for each of the combination is reported in Table~\ref{tbl:lin_order_res}.

\begin{table}[ht]
\normalsize
\centering
\begin{tabular}{+l^r^r^r}
               & Pre-order & Post-order & In-order \\
\midrule
LSTM 1 layer   & \textbf{34.14}     & 34.60      & 34.58    \\
LSTM 2-layer   & \textbf{33.69}     & 34.00      & 33.76    \\
biLSTM 1-layer & \textbf{34.46}     & 35.04      & 34.90    \\
biLSTM 2-layer & \textbf{33.88}     & 34.17      & 33.94    \\
CNN            & \textbf{36.54}     & 36.95      & 37.02    \\
\end{tabular}
\caption{\label{tbl:lin_order_res}
\textit{Lowest TER on development set for different models and linearization schemes}}
\end{table}

For all the models, the difference in performance between different linearization schemese is quite small.
However, across all models, the pre-order linearization is slightly better than the post-order and the in-order linearization.
Hence, for the subsequent experiments, we use pre-order linearization to convert from trees to sequences.

\subsection{Results}
\label{ssection:results}

Table~\ref{tbl:result} shows the prediction results by LSTM, biLSTM, and treeLSTM for three experimental scenarios listed in Table~\ref{tbl:ph_data}.
In scenario 1 and 2, biLSTM performed slightly worse than LSTM so we only compared LSTM against treeLSTM\@.
In scenario 1 where the training and test distributions were the same, treeLSTM yields 1.8\% $(p=2e^{-4})$ and 2.0\% $(p=6e^{-5})$ lower absolute TER (5.4\% and 6.0\% relative TER) than 1-layer and 2-layer LSTM respectively.
treeLSTM also yields 1.6\% $(p=0.06)$ and 0.6\% $(p=0.4)$ lower absolute SER (2.7\% and 1.0\% relative SER) than 1-layer and 2-layer LSTM respectively.
The trends are similar when individual output units (i.e., onset, nucleus, coda) are considered.
This result is unlikely due to treeLSTM having a higher capacity since the 2-layer LSTM had more parameters than treeLSTM.

\begin{table}[ht]
\normalsize
\centering
\begin{tabular}{+l^r^r^r^r^r}
               & SER  & TER  & On.  & Nu.  & Cd.  \\
\midrule
\multicolumn{6}{l}{Scenario 1: Tr, Sp $\rightarrow$ Tr, Sp} \\
\midrule
LSTM 1-layer   & 58.5 & 33.1 & 42.8 & 37.5 & 19.0 \\
LSTM 2-layer   & 57.5 & 33.3 & 42.8 & 38.3 & 18.9 \\
biLSTM 1-layer & 59.1 & 33.4 & 43.7 & 37.2 & 19.3 \\
biLSTM 2-layer & 57.8 & 32.9 & 42.5 & 36.9 & 19.2 \\
CNN            & 62.1 & 35.9 & 45.0 & 41.3 & 21.4 \\
\rowstyle{\bfseries}
treeLSTM       & 56.9 & 31.3 & 40.9 & 35.7 & 17.3 \\
\midrule
\multicolumn{6}{l}{Scenario 2: Non-Sp $\rightarrow$ Sp} \\
\midrule
LSTM 1-layer   & 73.5 & 48.5 & 57.3 & 53.0 & 35.3 \\
LSTM 2-layer   & 71.3 & 45.8 & 55.5 & 50.0 & 32.0 \\
biLSTM 1-layer & 74.1 & 48.4 & 57.2 & 53.0 & 35.0 \\
biLSTM 2-layer & 71.5 & 47.0 & 56.0 & 50.9 & 34.0 \\
CNN            & 79.1 & 52.1 & 62.4 & 56.9 & 37.1 \\
\rowstyle{\bfseries}
treeLSTM       & 69.6 & 43.8 & 51.8 & 48.6 & 31.0 \\
\midrule
\multicolumn{6}{l}{Scenario 3: Tr $\rightarrow$ Sp} \\
\midrule
LSTM 1-layer   & 77.2 & 55.5 & 62.2 & 59.5 & 44.8 \\
LSTM 2-layer   & 77.4 & 57.7 & 65.2 & 61.3 & 46.4 \\
biLSTM 1-layer & 73.5 & 51.6 & 57.9 & 55.2 & 41.8 \\
biLSTM 2-layer & 75.7 & 55.4 & 62.0 & 60.5 & 43.7 \\
CNN            & 70.5 & 48.1 & 54.1 & 49.7 & 40.5 \\
\rowstyle{\bfseries}
treeLSTM       & 68.8 & 47.7 & 53.7 & 50.7 & 38.9 \\
\end{tabular}
\caption{\label{tbl:result}
\textit{Cantonese phonemes prediction percentage error rate. Tr: Traditional, Sp: Simplified}}
\end{table}

When training and test distributions are different (scenario 2), models that have better inductive bias should perform better~\cite{haussler1988quantifying}.
For example, the convolution operation in convolutional neural network (CNN) has translation equivariant bias~\cite{Goodfellow-et-al-2016}.
This bias enforces that the representation of an object is the same regardless of its position in an image.
This bias makes CNN generalize much better and require few training samples than fully-connected neural networks.
For logographs, the inductive bias is that the interaction between sub-units is local in space.
This inductive bias is enforced in the treeLSTM model since a child node only interacts with its sibling.
The result is that the hierarchical embeddings is much more data-efficient than the LSTM\@.
The result shown in Table~\ref{tbl:result} indicates that treeLSTM can generalize better than LSTM models even when the test set has out-of-distribution samples.
treeLSTM yields 4.7\% $(p<1e^{-12})$ and 2.0\% $(p=6e^{-4})$ lower absolute TER (9.6\% and 4.3\% relative TER) than 1-layer LSTM and 2-layer LSTM respectively.
Besides, treeLSTM yields 3.9\% $(p=3e^{-6})$ and 1.7\% $(p=3e^{-2})$ lower absolute SER (5.3\% and 2.3\% relative SER) than 1-layer LSTM and 2-layer LSTM respectively.
The trends are similar when individual sub-syllabic classes (i.e., onset, nucleus, coda) are considered.

When training and test distributions are different and the amount of training data is limited, good inductive biases are even more important to obtain good generalization.
Comparing scenario 2 and 3, treeLSTM is less affected than LSTM by the limited training data.
In the limited training data regime, 2-layer LSTM clearly overfits badly compared to 1-layer LSTM and treeLSTM\@.
It is interesting to note that although the CNN model is the most competitive baseline in scenario 3 although it is worse than the LSTM and biLSTM when there is more data (scenario 1 and 2).
However, compared to the CNN, the treeLSTM still has lower SER $(p=0.07)$ and TER $(p=0.5)$\@.

\subsection{Ablation}
We conducted ablation experiments to see how much the models depend on the composition operators.
Without the operators, the LSTM, biLSTM, and CNN cannot discern the hierarchical grouping of sub-units.
On the other hand, even without the composition operations, the treeLSTM model still receives some structural information from ordering of the sub-units in a tree.

\begin{table}[ht]
\normalsize
\centering
\begin{tabular}{+l|r^r|r^r}
& \multicolumn{2}{c|}{+ operators} & \multicolumn{2}{c}{- operators}\\
\midrule
Model              & SER  & TER  & SER  & TER  \\
\midrule
LSTM 1-layer       & 58.5 & 33.1 & 62.0 & 35.5 \\
LSTM 2-layer       & 57.5 & 33.3 & 59.4 & 34.3 \\
biLSTM 1-layer     & 59.1 & 33.4 & 63.8 & 36.5 \\
biLSTM 2-layer     & 57.8 & 32.9 & 61.3 & 35.5 \\
CNN                & 62.1 & 35.9 & 67.2 & 40.2 \\
treeLSTM           & 56.9 & 31.3 & 57.3 & 32.0 \\
\end{tabular}
\caption{\label{tbl:ablation}
\textit{Results on the test set of scenario 1}}
\end{table}

In order to implement the case where there is no composition operators in the input,
for the LSTM, biLSTM, and CNN models, the operators were removed from the input sequences.
For the treeLSTM model, all the ${\bm V}$ terms were removed from the equation of the inner nodes.
We searched the best hyperparameters for each of the model using the development set of scenario 1.
We picked scenario 1 because it is the most common way to split data into training/validation/test sets, i.e. standard split.
The result is shown in Table~\ref{tbl:ablation}.
It can be seen that the composition operators do provide salient information for the task since taking them out results in worse performance across all models (as reflected by increases in error rates).
However, the treeLSTM performance does not degrade by much, it is more certain that treeLSTM learns to compose the sub-units chiefly from the tree structure.

\subsection{Prediction Order of Output Phonemes}
The phonetic subunits in Chinese characters usually predict nucleus and coda more reliably than onset.
This trend can be seen in Figure~\ref{fig:example} whereby all the nuclei and codas are the same across the first four characters which share the same phonetic subunit.
However, the most effective ordering of input and output in machine learning is may not align with human intuition.
For example, reversing the order of the input sentence boosted the performance of machine translation~\cite{sutskever2014sequence}, while swapping the order of onsets and nuclei in Thai syllables boosted the performance of English-to-Thai transliteration~\cite{nguyen2016regulating}.
We adopted the Coda-Nucleus-Onset prediction order in this paper as shown in Section~\ref{ssec:task_layer}.
However, we also tried using a different prediction order which is Onset-Nucleus-Coda.
We replaced the task-specific layer of the proposed model and searched for the optimal hyperparameters.
The model with the best hyperparameters is then applied on the test set.
Empirically, we observed little difference in performance between the two orders.

\begin{table}[ht]
\normalsize
\centering
\begin{tabular}{+l^r^r^r^r^r}
Output order       & SER  & TER  & On.  & Nu.  & Cd.  \\
\midrule
Coda-Nucleus-Onset & 56.9 & 31.3 & 40.9 & 35.7 & 17.3 \\
Onset-Nucleus-Coda & 57.3 & 32.0 & 42.1 & 36.8 & 17.2 \\
\end{tabular}
\caption{\label{tbl:out_order}
\textit{Comparing different orders of predicting output phonemes. treeLSTM results on the test set of scenario 1}}
\end{table}

\section{Experiments --- Language Modeling}
\label{sec:exp_lm}

We evaluated how well the hierarchical embeddings can improve language modeling in Chinese.
We compare hierarchical embeddings against standard embeddings to quantify the usefulness of sub-unit semantic information since hierarchical embeddings are imbued with semantic information from the sub-units while standard embeddings are not. 

\subsection{Data}

As the characters (logographs) in the output of language models are not independent, it is difficult to design meaningful statistical tests to evaluate the effectiveness of our proposed approach.
Instead we chose a wide variety of five different datasets, consisting of three datasets using simplified characters (Chinese Penn Treebank (CTB) Version 5.1~\cite{xue2005penn}, Beijing University (PKU) dataset~\cite{emerson2005second}, and Microsoft Research (MSR) dataset~\cite{emerson2005second}) and two datasets using traditional characters (City University of Hong Kong (CITYU) dataset~\cite{emerson2005second} and Academia Sinica (AS) dataset~\cite{emerson2005second}).
If we can show consistent improvements across these datasets, it implies the proposed hierarchical embeddings are effective.
Table~\ref{tbl:lm_data} shows the data split for each of the datasets.
Data splits for CTB and PKU datasets are taken from~\cite{kawakami2018unsupervised}.\footnote{\url{https://s3.eu-west-2.amazonaws.com/k-kawakami/seg.zip}}

\begin{table}[ht]
\normalsize
\centering
\begin{tabular}{+l^r^r^r^r}
    Dataset & Training & Validation & Test \\
\midrule
    CTB (Simplified) & 50,734 & 349 & 345 \\
    PKU (Simplified) & 17,149 & 1,841 & 1,790 \\
    MSR (Simplified) & 83,000 & 3,924 & 3,985 \\
    CITYU (Traditional) & 51,000 & 2,019 & 1,493 \\
    AS (Traditional) & 690,000 & 18,953 & 14,431 \\
\end{tabular}
\caption{\label{tbl:lm_data}
\textit{Number of sentences in the training, validation, and test sets in each of the datasets.}}
\end{table}

\subsection{Setup}

\begin{figure}[ht]
\centering
\begin{subfigure}{\columnwidth}
\centering
\includegraphics[width=\linewidth]{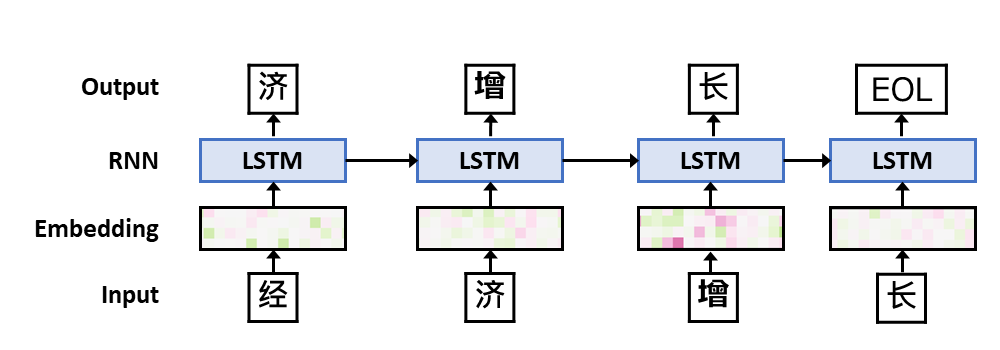}
\caption{\label{fig:lm_model_bl}
    \textit{Standard embeddings (baseline)
    }}
\end{subfigure}

\begin{subfigure}{\columnwidth}
\centering
\includegraphics[width=\linewidth]{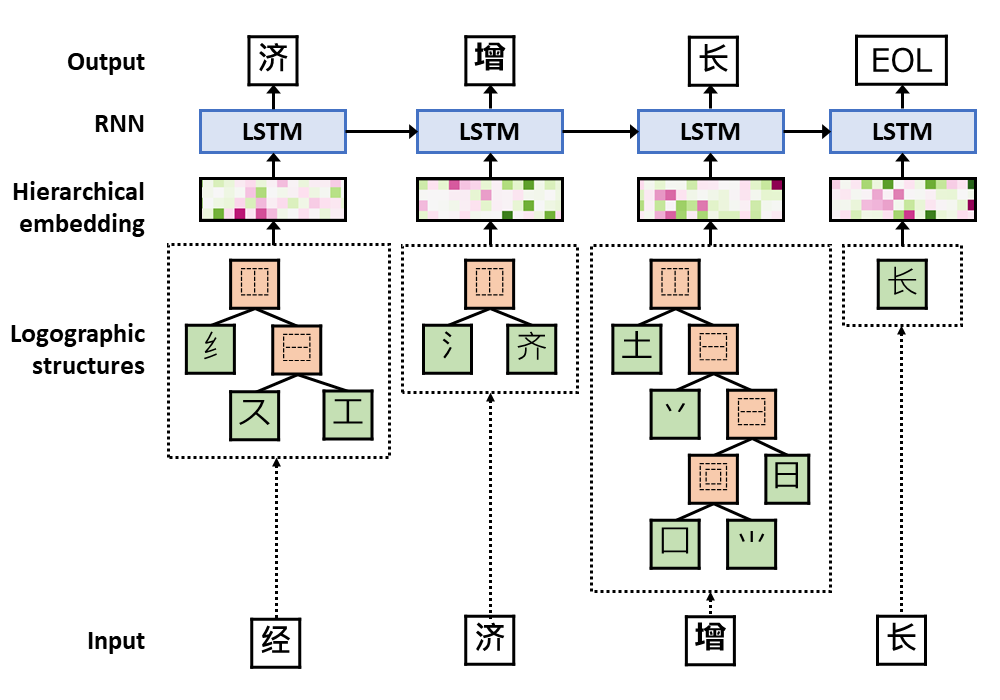}
\caption{\label{fig:lm_model_hier}
    \textit{Hierarchical embeddings (proposed)
    }}
\end{subfigure}
\caption{\textit{Language model (LM)}}
\end{figure}

We used AWD-LSTM (ASGD Weight-Dropped LSTM) model~\cite{merity2017regularizing} as the core in the language modeling experiment.
The input to AWD-LSTM is either hierarchical embeddings (Figure~\ref{fig:lm_model_hier}) or standard character embeddings (Figure~\ref{fig:lm_model_bl}).
We considered the standard character embeddings as the baseline.
We trained the model using the training set for a fixed number of epochs and used the validation set to select the best model.
The best model performance was evaluated on the test set after training finished.

\subsection{Metrics}

We evaluated models' performance using perplexity (PPL) and bits-per-character (BPC).
BPC is a standard evaluation metric for character-level LMs~\cite{graves2013generating}.

\begin{eqnarray}
    BPC & = & - \frac{1}{|\bm{x}|}\sum{\log_{2}{p(x_t|\bm{x}_{<t})}} \nonumber\\
    PPL & = & 2^{BPC} \nonumber
\end{eqnarray}
where $\bm{x}$ is the whole corpus, $x_t$ is the character at position $t$, and $|\bm{x}|$ is the length of the corpus.

\subsection{Hyperparameters}

The same hyperparameters are used across the datasets.
We optimized the models using the Adam~\cite{kingma2015adam} optimizer for 300 epochs.
The learning rate was set at 0.002 and is divided by 10 after 250 epochs.
The size of hidden layer is fixed as 1000.
The size of the embedding is fixed as 200.
The AWD-LSTM has three hidden layers with sizes 1000, 1000, 200 respectively.
We used dropout~\cite{srivastava14dropout} on input and hidden layers to prevent overfitting.
Dropout rates were set as 0.1, 0.1, and 0.25 for the input, hidden and output layers of the AWD-LSTM.
L2 weight decay was set as $1.2\times10^{-6}$.
Weight dropout was set at 0.5.
The batch size was 100.
To improve computational speed, only embeddings of the characters appearing in the training batch were updated.
During testing, the embeddings were constructed once and then cached, hence using hierarchical embedding was nearly as fast as standard embeddings.
The caching technique was similar to~\cite{ling2015finding}.

\subsection{Results}
\label{ssection:results}

\begin{table}[ht]
\normalsize
\centering
\begin{tabular}{+l^r^r}
    Model      & Perplexity  & BPC  \\
\midrule
\multicolumn{3}{l}{Dataset: CTB (Simplified)} \\
\midrule
LSTM~\cite{kawakami2018unsupervised} & 30.78 & 4.944 \\
Segmental Neural LM~\cite{kawakami2018unsupervised} & 28.56 & 4.836 \\
AWD-LSTM, baseline & 19.14 & 4.259 \\
\rowstyle{\bfseries}
AWD-LSTM, hier-emb & 18.71 & 4.226 \\
AWD-LSTM, hier-emb, ext & 18.85 & 4.237 \\
\midrule
\multicolumn{3}{l}{Dataset: PKU (Simplified)} \\
\midrule
LSTM~\cite{kawakami2018unsupervised} & 73.66 & 6.203 \\
Segmental Neural LM~\cite{kawakami2018unsupervised} & 59.01 & 5.883 \\
AWD-LSTM, baseline & 55.42 & 5.792 \\
\rowstyle{\bfseries}
AWD-LSTM, hier-emb & 53.96 & 5.754 \\
AWD-LSTM, hier-emb, ext & 56.09 & 5.810 \\
\midrule
\multicolumn{3}{l}{Dataset: MSR (Simplified)} \\
\midrule
GRU~\cite{dai2017glyph} & 47.53 & 5.571 \\
GRU, glyph-emb~\cite{dai2017glyph} & 47.75 & 5.577 \\
GRU, reimplemented & 34.27 & 5.099 \\
GRU, glyph-emb, reimplemented & 34.76 & 5.119 \\
AWD-LSTM, baseline & 22.28 & 4.478 \\
AWD-LSTM, glyph-emb & 22.52 & 4.493 \\
AWD-LSTM, hier-emb & 22.64 & 4.501 \\
\rowstyle{\bfseries}
AWD-LSTM, hier-emb, ext & 22.25 & 4.476 \\
\midrule
\multicolumn{3}{l}{Dataset: CITYU (Traditional)} \\
\midrule
AWD-LSTM, baseline & 70.48 & 6.139 \\
\rowstyle{\bfseries}
AWD-LSTM, hier-emb & 68.47 & 6.097 \\
AWD-LSTM, hier-emb, ext & 68.93 & 6.107 \\
\midrule
\multicolumn{3}{l}{Dataset: AS (Traditional)} \\
\midrule
AWD-LSTM, baseline & 45.99 & 5.523 \\
AWD-LSTM, hier-emb & 46.88 & 5.551 \\
\rowstyle{\bfseries}
AWD-LSTM, hier-emb, ext & 45.91 & 5.521 \\
\end{tabular}
\caption{\label{tbl:lm_result}
\textit{Language modeling performance on test sets from different datasets.
\textbf{hier-emb}: hierarchical embedding, \textbf{glyph-emb}:  glyph embeddings, \textbf{baseline}: standard embeddings, \textbf{ext}: additional bias term in treeLSTM. Results for the LSTM, Segmental Neural LM, and the glyph embeddings were taken from the original papers.
We also reimplemented the glyph embeddings for a fairer comparison.
}}
\end{table}

Table~\ref{tbl:lm_result} shows the prediction results.
We also report results on the CTB and PKU datasets from~\cite{kawakami2018unsupervised} and results on the MSR dataset from~\cite{dai2017glyph}.
The results from~\cite{kawakami2018unsupervised} can be compared with our results since the results are evaluated on the same data splits.
However, direct comparison is unfair for~\cite{kawakami2018unsupervised} because our models are bigger than theirs.
The results from~\cite{dai2017glyph} cannot be compared with our results as the data splits are different because their data split is not publicly available.
Thus, we reimplemented the glyph embedding for a fairer comparison.
The glyph embedding model architecture is similar to that used in the original paper~\cite{dai2017glyph}.
We only include the results for the Segmental Neural LM model for reference and did not reimplement this model because it depends on multitask training which is different from the other models.
Our result agrees with the conclusion from~\cite{dai2017glyph} that the glyph embeddings are slightly worse than standard embeddings regardless of the baseline (GRU or AWD-LSTM).
The hierarchical embeddings outperformed the standard embeddings in all datasets, regardless of whether the datasets use simplified or traditional characters.

\section{Relation to Other Work}
\label{sec:related}

\subsection{Exploiting Recursive Structures}

Exploiting recursive structures has been shown to be beneficial in many NLP tasks such as sentiment analysis~\cite{irsoy2014deep,tai2015improved,zhu2015long}, text simplification~\cite{siddharthan2014hybrid}, and machine translation~\cite{quirk2005dependency,eriguchi2016tree,nakazawa2016insertion,chen2017improved}.
These models are usually trained using human annotated structures but may be tested on structures annotated automatically using parsers when human annotation is not available.
This mismatch in annotation quality could worsen the performance of these models and could partially explain why exploiting structures in NLP tasks have not always led to better results.
For example, recursive models~\cite{socher2011semi,socher2012semantic,socher2013recursive} were not as good as the biLSTM in sentiment analysis task~\cite{tai2015improved}.
To address the mismatch in annotation quality, new models which can both produce and exploit structures have been introduced~\cite{bowman2016fast,eriguchi2017learning,yogatama2017learning}.
For our case, annotation quality is consistent across the training and test set, thus, better ways of modeling structures led to better results.

\subsection{Building Logographic Embeddings}

In languages like Mandarin, Japanese or Cantonese, logographs are characters and the number of characters are in the range of thousands. 
In contrast, alphabetic languages usually have far fewer characters (e.g. 26 characters for English).
The large number of characters in languages with logographic origin makes character-level modeling inefficient and worsens the problem of out-of-vocabulary words and characters.
However, alphabetic languages and languages with logographic origin are often treated the same way, disregarding their intrinsically marked differences~\cite{zhang2018neural}. 
Modeling logograph sub-units can alleviate these issues since there are fewer sub-units and they can be used to construct out-of-vocabulary words and characters.
This is consistent with how learners of languages with logographic origin can comprehend the meaning or pronunciation of a logograph from its constituent sub-units~\cite{ho1997phonological}.
Hence, leveraging structures of logographs can be useful in capturing semantic~\cite{su2017learning,song2018joint} or phonological information~\cite{nguyen2018multimodal}.

There are many prior work on building embeddings of logographs.
The first approach is to apply convolutional neural network (CNN) on the visual rendering of logographs~\cite{dai2017glyph,liu2017learning,toyama2017utilizing,su2017learning}.
The second approach is to combine sub-unit embeddings with the logograph embeddings.
Sub-units embeddings can be learned independently of logograph embeddings~\cite{shi2015radical,peng2017radical,yu2017joint} using Skip-Gram or CBOW models~\cite{mikolov2013distributed} or learned jointly with logograph embeddings~\cite{yin2016multi,ke2017radical,karpinska2018subcharacter}.
The third approach is to apply CNN or RNN on the sequence of sub-units~\cite{dong2016character,han2017dual,zhuang2017natural,cao2018cw2vec,li2018subword}.

Our work is most similar to the third approach.
However, while our approach exploits the recursive structures of logographs, most work in this area ignores structures or only consider the structures implicitly.

\subsection{Incorporating Morphology into Embeddings}

In languages like English, popular models to learn word embeddings assign a distinct vector to each word, ignoring word morphology (how characters, word's sub-units, form a word).
This approach uses solely the context surrounding words to learn the embeddings which may be a limitation in languages with a large vocabulary and many rare words since the context may be insufficient to learn good embeddings.
Building logographic (character) embeddings in languages of logographic origin has the same difficulty since there a lot of logographs (characters) and many of them are rare characters.

To incorporate morphology into word embeddings learning,~\cite{bojanowski2017enriching,zhao2018generalizing} proposed building word embeddings by averaging bags of character n-grams.
This method may be agnostic to the order of characters if the n-gram length is short.
Others have used RNN~\cite{ling2015finding,li2018subword} or CNN~\cite{kim2016character,papay2018addressing,li2018subword} to better incorporate word morphology information into words embeddings.
Unlike English words which are linear sequences of characters, logographs are recursive structures of of sub-units.
Hence, using models operating on sequences such as RNN or LSTM may not be optimal.

Rare word/character embeddings can be improved by leveraging similarity in morphology between rare words and common words.
\cite{pinter2017mimicking,kim2018learning,schick2019attentive} proposed building embeddings of new words from pre-trained embeddings by learning mapping from characters to embeddings.
However, in this line of approach, the embeddings are fixed, which may not be useful for tasks that require information not captured in the embeddings pre-trained via unsupervised language modeling.
In work from~\cite{ling2015finding,kim2016character,li2018subword}, the embeddings are learned jointly with the task models so that the embeddings contain useful information for the task.
Our hierarchical embeddings can be trained on task-specific data, making it potentially useful for many different tasks.

\section{Discussion}
\label{sec:discussion}

\begin{figure*}[ht]
\centering
\begin{subfigure}{.5\textwidth}
  \centering
  \includegraphics[width=.95\linewidth]{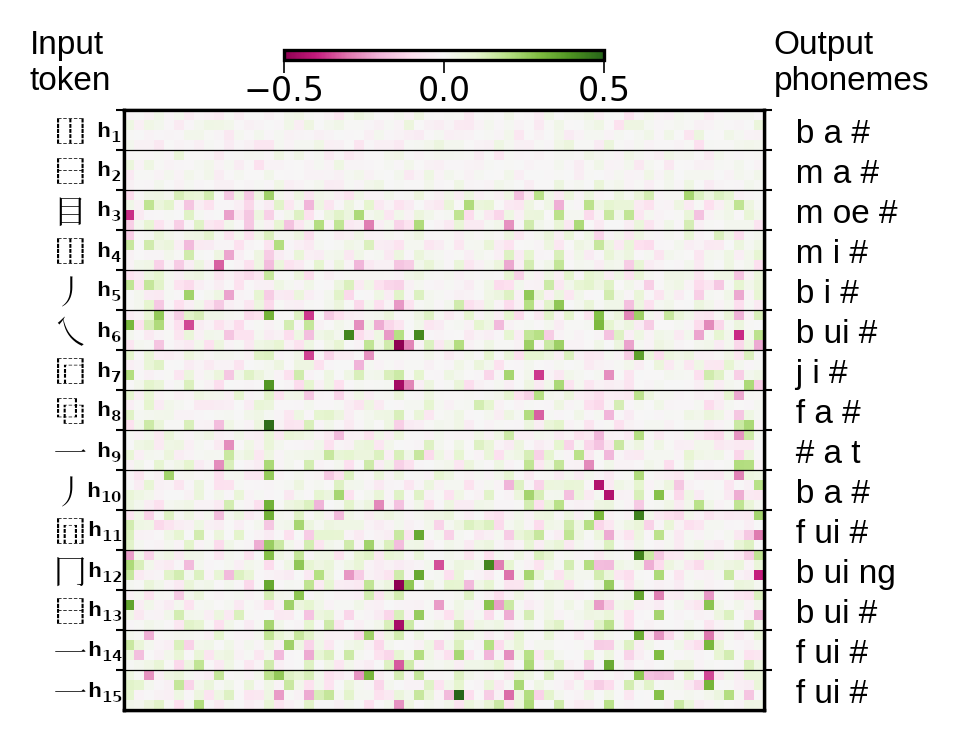}
  \caption{\label{subfig:ex1_slstm}LSTM prediction}
\end{subfigure}%
\begin{subfigure}{.5\textwidth}
  \centering
  \includegraphics[width=.95\linewidth]{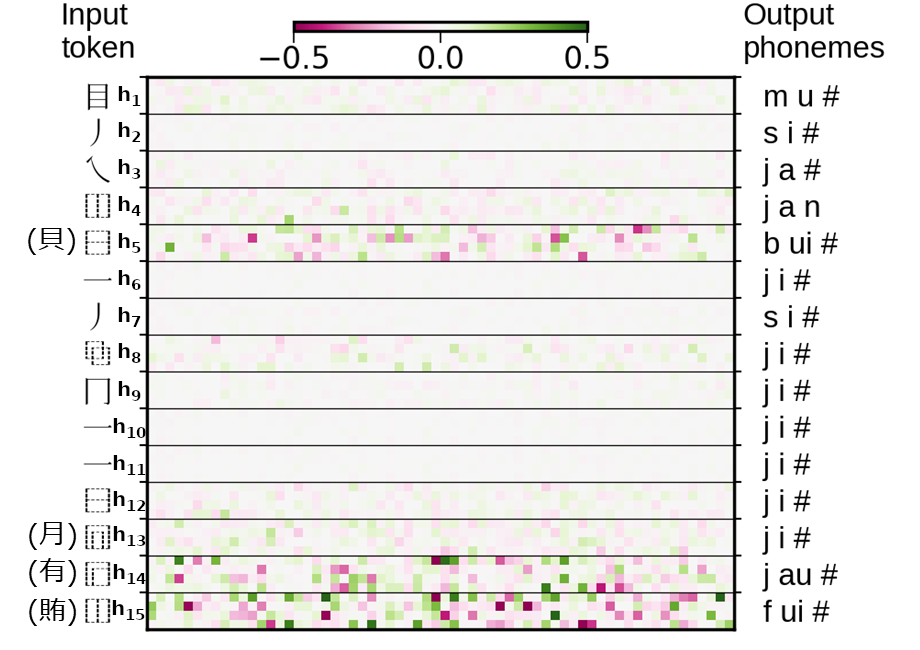}
  \caption{\label{subfig:ex1_tlstm}treeLSTM prediction}
\end{subfigure}
\caption{\label{subfig:ex1}
\textit{Visualizing the construction of the logograph embedding for
\protect\begin{CJK*}{UTF8}{bkai}賄\protect\end{CJK*} (\emph{bribery}) by LSTM (a) and treeLSTM (b).
The central panels show the hidden states $\textbf{h}_i$.
The left columns show the input sub-units.
The right columns show the predicted pronunciations using the hidden states $\textbf{h}_i$.
The bottom rows of the right columns are the predicted pronunciations for the logographs (``f ui \#'' for both LSTM and treeLSTM).
Ground-truth pronunciation is ``f ui \#''.
}}
\end{figure*}

\begin{figure*}[hb]
\centering
\begin{subfigure}{.5\textwidth}
\centering
\includegraphics[width=.95\linewidth]{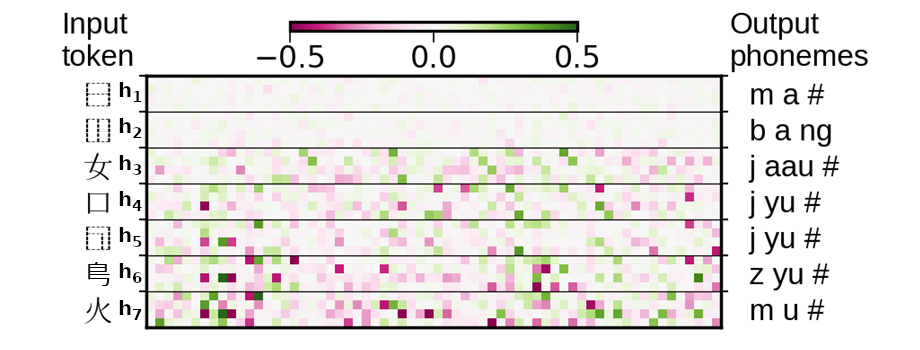}
\caption{\label{subfig:ex2_slstm}LSTM prediction}
\end{subfigure}%
\begin{subfigure}{.5\textwidth}
\centering
\includegraphics[width=.95\linewidth]{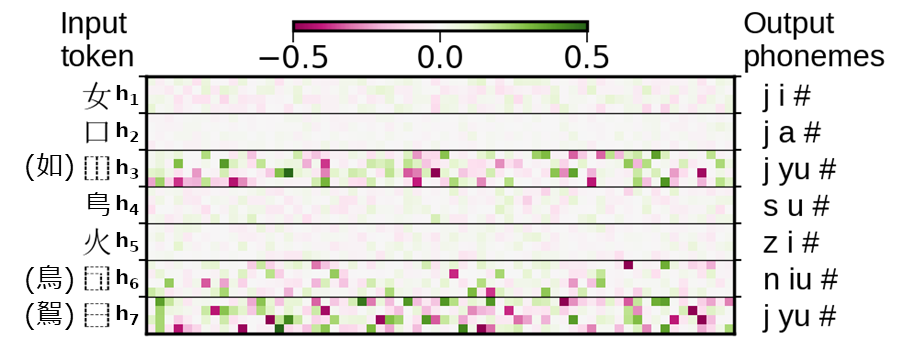}
\caption{\label{subfig:ex2_tlstm}treeLSTM prediction}
\end{subfigure}
\caption{\label{subfig:ex2}
\textit{Visualizing the construction of the logograph embedding for
\protect\begin{CJK*}{UTF8}{bkai}鴽\protect\end{CJK*} (\emph{quail}) by LSTM (a) and treeLSTM (b).
The central panels show the hidden states $\textbf{h}_i$.
The left columns show the input sub-units.
The right columns show the predicted pronunciations using the hidden states $\textbf{h}_i$.
The bottom rows of the right columns are the predicted pronunciations for the logographs (``m u \#'' for LSTM and ``j yu \#'' for treeLSTM).
Ground-truth pronunciation is ``j yu \#''.
While LSTM made a mistake, treeLSTM predicted the correct pronunciation.
}}
\end{figure*}

\subsection{Left-right Bias in Pronunciation Prediction}

More than 80\% of frequently used Han logographs are semantic-phonetic compounds~\cite{li1993analysis}.
These compounds consist of sub-units that might contain phonetic or semantic information~\cite{hsiao2006analysis}.
Pronunciation of these compounds could conceivably be predicted from the phonetic sub-units.
Amongst semantic-phonetic compounds, logographs with the left-right arrangement (in which the semantic sub-unit is on the left and the phonetic sub-unit is on the right) are the most common.
For logographs with the left-right arrangement, a good model for logograph's pronunciation should prefer the right child (the likely phonetic sub-unit) of a root node for making pronunciation prediction.
To check whether the hierarchical embeddings prefer the left child or the right child, we compared the norm of the left forget gate against the norm of the right forget gate.
The right child is preferred if the norm of the right forget gate is larger.

\begin{table}[ht]
\normalsize
\centering
\begin{tabular}{+l@{\hspace{1em}}^c@{\hspace{1em}}^c}
    Scenario  & Left-Right & Prefer Right  \\
\midrule
    Tr., Sp. $\rightarrow$ Tr., Sp. & 1657 & 1543 (93\%) \\
    Non-Sp. $\rightarrow$ Sp.  & 1686 & 1589 (94\%) \\
    Tr. $\rightarrow$ Sp. & 1686 & 1643 (97\%) \\
\end{tabular}
\caption{\label{tbl:lr_bias}
\textit{Number of times the model using hierarchical embeddings predicts the phonetic sub-unit is on the right of a logograph that follows the left-right arrangement.
    The scenarios were described in Table~\ref{tbl:ph_data}.
    Tr: Traditional, Sp: Simplified.
    }}
\end{table}

In Table~\ref{tbl:lr_bias}, the second column shows the number of logographs following the left-right arrangement for different scenarios.
The third column shows the number of logographs following the left-right arrangement in which the right child is preferred over the left child.
The hierarchical embeddings prefer the right child most of the time (close to 100\%) in all three scenarios.
Thus, the learned hierarchical embeddings consider the right sub-units to be more relevant for pronunciation prediction for the majority of compound logographs with the left-right arrangement.
This is consistent with human intuition.
Since human depends on this intuition to infer pronunciation and it seems to work well, this suggests that the hierarchical embeddings might have learned a general solution that works well.

\subsection{Robustness to Distractors in Pronunciation Prediction}

By overfitting to common patterns at the expense of more difficult, infrequent samples that require deeper understanding, statistical models can perform well as measured by some aggregate metrics~\cite{jia2017adversarial}.
A common pattern useful for predicting pronunciation is that phonetic sub-units usually occur at the end of the linearized sequences.
A general model would be able to find where the phonetic sub-units are in the sequences.
A model that only attends to the end of sequences would make wrong prediction when the phonetic sub-units are not at the end of the sequences.

To determine how the models predict, we visualize the hidden states of LSTM and treeLSTM\@.
The visualization for biLSTM is not shown since it performed worse than LSTM\@.
For both LSTM and treeLSTM, the last hidden state (e.g. $\textbf{h}_{15}$ in Figure~\ref{subfig:ex1}) is considered the logograph embedding.
The intermediate embeddings (e.g. $\textbf{h}_1$ to $\textbf{h}_{14}$) are embeddings of the subsequence of sub-units for LSTM and
embeddings of the subtrees of sub-units for treeLSTM\@.
The hidden states (embeddings) evolve to contain more phonetic information with more sub-units as indicated by generally increasing magnitude of the hidden states (corresponding to darker bands).
When the magnitude of the hidden states are small (corresponding to faint bands), the hidden states do not have enough information to predict pronunciation confidently.
We also obtained the prediction corresponding to each hidden state by feeding the hidden states ($\textbf{h}_1$ to $\textbf{h}_{15}$) to the task-specific layer in order to determine at which step did the embeddings contain phonetic information to make the correct pronunciation prediction.

Figure~\ref{subfig:ex1} shows how the models predict the pronunciation of the logograph \begin{CJK*}{UTF8}{bkai}賄\end{CJK*} (\emph{bribery}).
This is a common example as the phonetic sub-units are on the right (corresponding to end of the linearized sequence).
While both models predict correctly, they used the logograph structural representation differently.
LSTM had to observe the whole sequence to predict correctly, as suggested by the build-up in magnitude of the embeddings until the end of the sequence.
For treeLSTM, the pattern of the embeddings' magnitude is consistent with the hierarchical structure of the input logograph with two subtrees \begin{CJK*}{UTF8}{bkai}貝\end{CJK*} and \begin{CJK*}{UTF8}{bkai}有\end{CJK*}.
Specifically, not only was the final pronunciation prediction of \begin{CJK*}{UTF8}{bkai}賄\end{CJK*} correct (``f ui \#''), but pronunciation of the subtrees (\begin{CJK*}{UTF8}{bkai}貝\end{CJK*} and \begin{CJK*}{UTF8}{bkai}有\end{CJK*}) were also correct (``b ui \#'' and ``j au \#'' respectively).

Figure~\ref{subfig:ex2} shows a rare example where the phonetic sub-units are not at the end of the sequence.
LSTM made the correct prediction after observing the relevant parts (up to the second last input token) but soon forgot the correct prediction as it might focus more on the end of the sequence.
This mistake indicates that LSTM might have learned a heuristic instead of the general strategies.
On the other hand, treeLSTM predicted the pronunciation correctly by seemingly focusing on the relevant part (\begin{CJK*}{UTF8}{bkai}如\end{CJK*}) of the logograph and ignoring the less relevant tokens.
Thus, imposing a prior on the mapping from logographs to embeddings by using recursive network seems to lead to a solution that may generalize better to more challenging cases.

\subsection{Infrequent Characters' Embeddings in Language Modeling}

\begin{table}[ht]
\centering
\begin{tabular}{+l|l|l}
Character & Standard Embedding & Hierarchical Embedding\\
\midrule
\includegraphics[width=.04\linewidth]{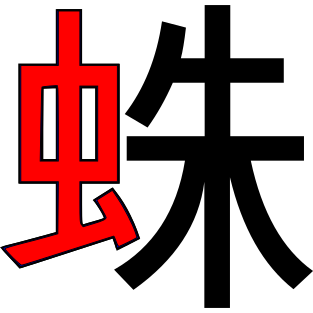},
& \includegraphics[width=.04\linewidth]{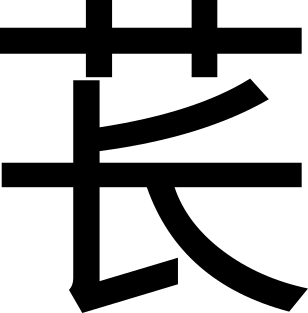}, \emph{a plant}, ``ch a ng''
& \includegraphics[width=.04\linewidth]{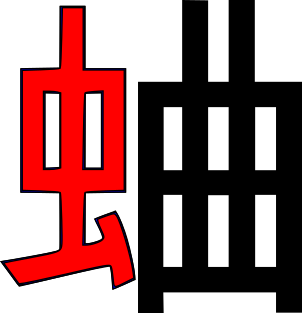}, \emph{cricket}, ``q u \#'' \\
\emph{spider},
& \includegraphics[width=.04\linewidth]{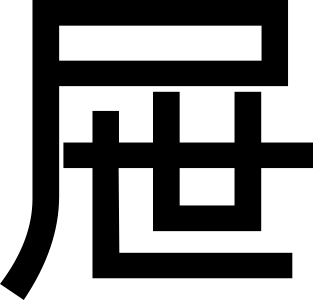}, \emph{drawer}, ``t i \#''
& \includegraphics[width=.04\linewidth]{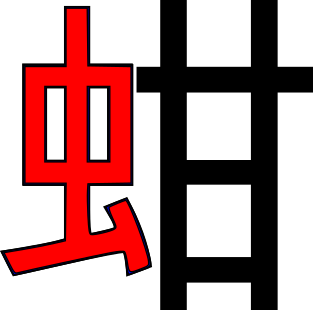}, \emph{ark clam}, ``q u \#'' \\
``zh u \#''
& \includegraphics[width=.04\linewidth]{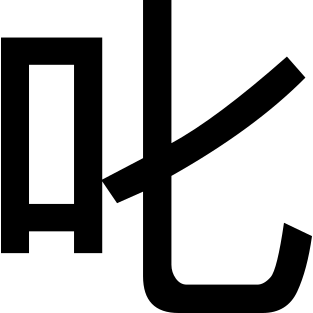}, \emph{scold}, ``ch i \#''
& \includegraphics[width=.04\linewidth]{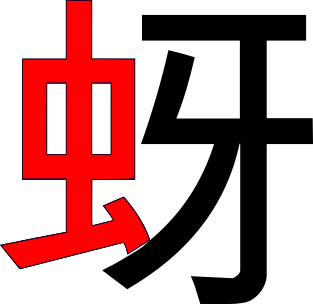}, \emph{louse}, ``y a \#'' \\
\midrule
\includegraphics[width=.04\linewidth]{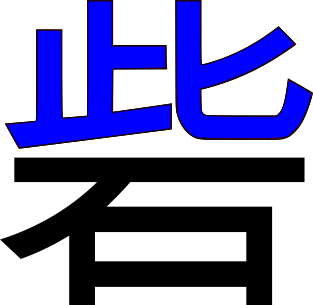},
& \includegraphics[width=.04\linewidth]{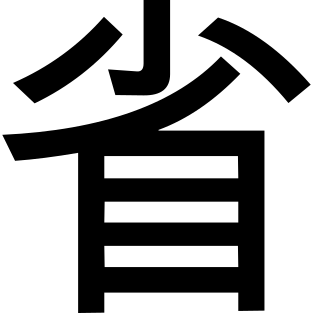}, \emph{omit}, ``sh e ng''
& \includegraphics[width=.04\linewidth]{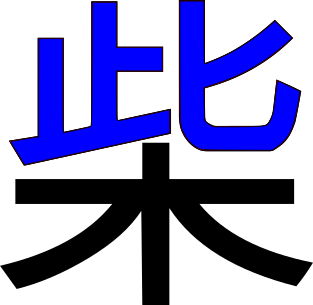}, \emph{firewood}, ``ch ai \#''\\
\emph{fort},
& \includegraphics[width=.04\linewidth]{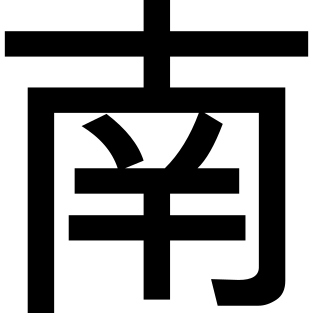}, \emph{south}, ``n a n''
& \includegraphics[width=.04\linewidth]{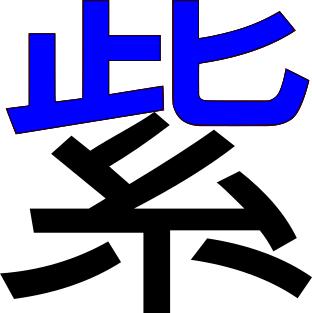}, \emph{purple}, ``z i \#'' \\
``zh ai \#''
& \includegraphics[width=.04\linewidth]{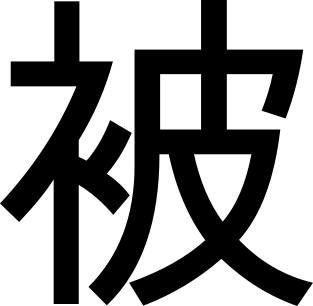}, \emph{blanket}, ``b ei \#''
& \includegraphics[width=.04\linewidth]{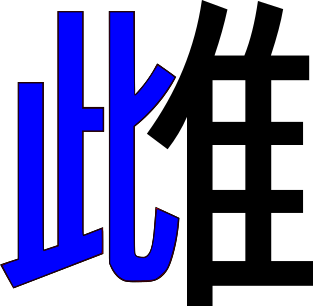}, \emph{female}, ``c i \#'' \\
\midrule
\includegraphics[width=.04\linewidth]{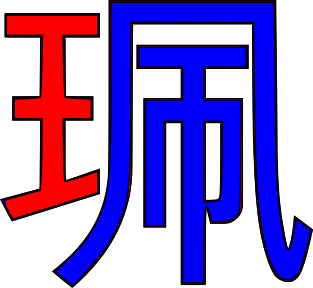},
& \includegraphics[width=.04\linewidth]{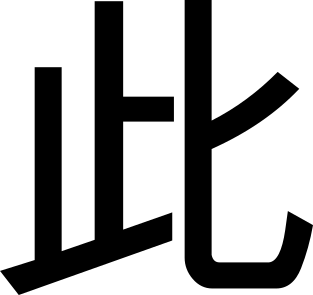}, \emph{this}, ``c i \#''
& \includegraphics[width=.04\linewidth]{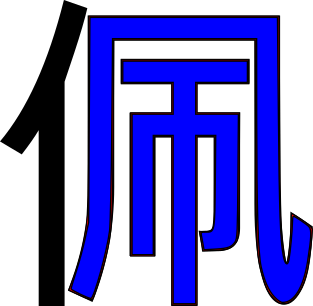}, \emph{pendant}, ``p ei \#'' \\
\emph{jade belt},
& \includegraphics[width=.04\linewidth]{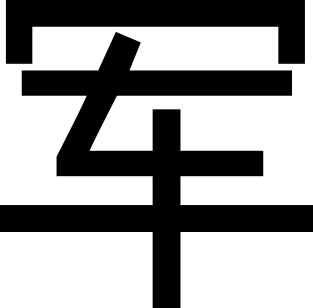}, \emph{army}, ``j u n''
& \includegraphics[width=.04\linewidth]{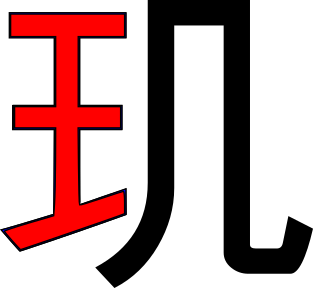}, \emph{imperfect pearl}, ``j i \#'' \\
``p ei \#''
& \includegraphics[width=.04\linewidth]{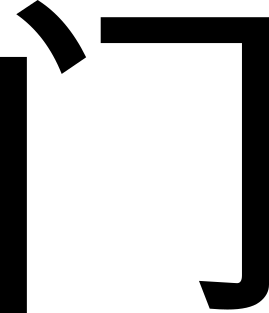}, \emph{gate}, ``m e n''
& \includegraphics[width=.04\linewidth]{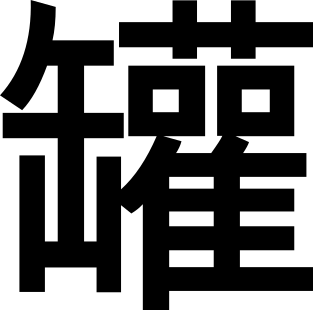}, \emph{watering can}, ``g ua n'' \\
\midrule
\includegraphics[width=.04\linewidth]{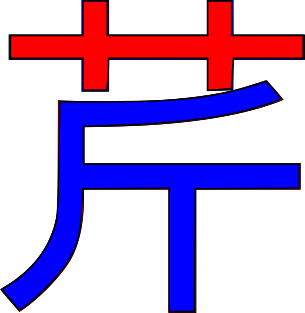},
& \includegraphics[width=.04\linewidth]{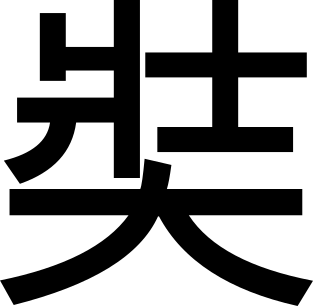}, \emph{powerful}, ``z a ng''
& \includegraphics[width=.04\linewidth]{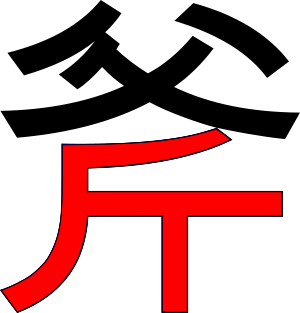}, \emph{axe}, ``f u \#''\\
\emph{celery},
& \includegraphics[width=.04\linewidth]{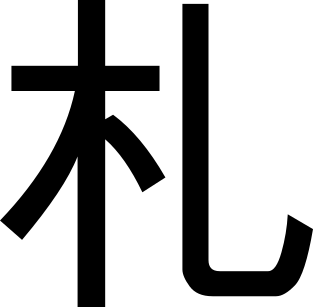}, \emph{note}, ``zh a \#''
& \includegraphics[width=.04\linewidth]{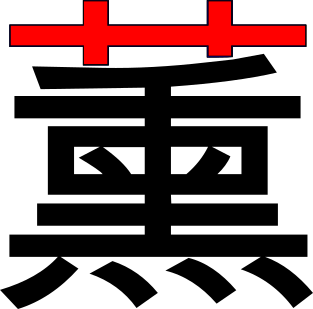}, \emph{fragrance}, ``x u n''\\
``q i n''
& \includegraphics[width=.04\linewidth]{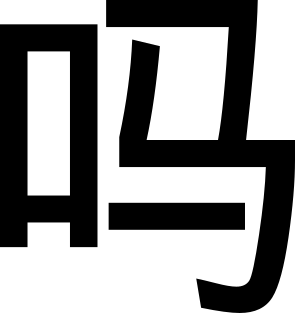}, \emph{not}, ``m a \#''
& \includegraphics[width=.04\linewidth]{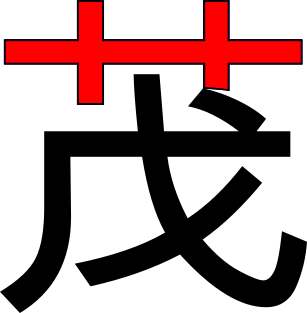}, \emph{lush}, ``m ao \#''\\
\end{tabular}
\caption{\label{tbl:lm_rare}
\textit{Nearest neighbors in embedding space of infrequent words.
The meaning and Mandarin pronunciation are shown next to the characters.
The common sub-units between the logographs and their neighbors in the embedding space are color-coded.
Red sub-units carry semantic information. Blue sub-units carry phonetic information.
}}
\end{table}

Hierarchical embeddings could learn better representations of infrequent characters than standard embeddings could since the latter ignores the morphology within characters.
Using the learned embeddings in the language modeling experiments, we looked for characters that are most similar (nearest neighbors) to the infrequent characters in the embedding space.
If the nearest neighbors are semantically or phonologically close then we are more certain that the learned embeddings are sensible.
The distance between embedding vectors is calculated using cosine similarity.
Table~\ref{tbl:lm_rare} showed that the infrequent characters and their nearest neighbors are relatively close in meaning when using hierarchical embeddings.

For standard embeddings, infrequent characters and their neighbors are generally unrelated.
For example, \emph{spider} are unrelated to \emph{a plant}, \emph{drawer}, or \emph{scold}.
It is possible that with little training data, the infrequent characters' embedding stay close to the original random initialized values and hence are far away from related characters in embedding space.
For hierarchical embeddings, infrequent characters are more related to their neighbors.
The relatedness between infrequent characters and their neighbors can be semantic or phonological.
For example, the first row in Table~\ref{tbl:lm_rare} shows characters (\begin{CJK*}{UTF8}{bkai}{\Large蛛,蛐,蚶,蚜}\end{CJK*}) that share the same sematic sub-units (shown in red).
Accordingly, \emph{spider} is semantically related to \emph{cricket}, \emph{ark clam}, and \emph{louse} since they are all insects.
The second row in Table~\ref{tbl:lm_rare} shows another example in which characters (\begin{CJK*}{UTF8}{bkai}{\Large砦,柴,紫,雌}\end{CJK*}) have the same phonetic sub-units (shown in blue).
Correspondingly, ``zh ai \#'' is phonologically related (having similar pronunciation) to  ``ch ai \#'', ``z i \#'', and ``c i \#''.

However, the hierarchical embeddings are not always accurate and the last row of Table~\ref{tbl:lm_rare} shows an interesting failure.
The cosine distance suggests that (\emph{celery}, ``q i n'') and (\emph{axe}, ``f u \#'') are related.
Although both characters has a common sub-unit, the sub-unit carries phonological information (color-coded as blue) in the case of  (\emph{celery}, ``q i n'') while the sub-unit carries semantic information (color-coded as red) in the case of (\emph{axe}, ``f u \#'').

\subsection{Automated Feature Granularities Selection}

The granularity of the input features derived from logographs could have a major impact on model performance.
The input features could be as granular as individual strokes, which results in a small vocabulary.
Different permutations of the strokes can form unique ideographs and expand the vocabulary.
The choice of the vocabulary set has a major impact on sequential models like RNN, as a big vocabulary makes training slow and makes it hard for the model to generalize.
On the other hand, a small vocabulary leads to longer sequences and makes it harder for models to learn.
\cite{nguyen2017sub} showed that a big vocabulary yields lower perplexity for language modeling of Japanese, while big vocabulary implies that each token is a meaningful unit that carries semantic information~\cite{karpinska2018subcharacter}.
Moreover, different sub-unit granularities might be more suitable for different logographic languages.
For example, ideographs are more suitable as input tokens for Chinese, while individual strokes are better suited for Japanese~\cite{zhang2018neural}.
\cite{su2017learning} chose to extract visual features of logographs instead of symbolic features to avoid specifying the level of granularity when decomposing a character.

In Figure~\ref{subfig:ex1_slstm} and~\ref{subfig:ex2_slstm}, LSTM treats all input features with relatively equal importance, evident by relatively high activation values across most hidden states.
On the contrary, the structural constraints imposed by treeLSTM resulted in a more automated selection of input features, in which most of the high activation concentrate at the hidden states of sub-trees' roots.
In other words, treeLSTM seemed to have learned to build representations relevant to the task at the right level of granularity.
Learning the right features via structures instead of delicate feature engineering is an advantage that should be explored further for RNN models.

\subsection{Intuitive Exploitation of Input Structures}

Various work suggested that incorporating syntactic structures is tricky and does not always improve results.
For example, in subject-verb agreement modeling, a model could easily ignore syntactic information from the input data and so syntactic constraints must be explicitly injected into the model's architecture~\cite{kuncoro2018lstms}.
Doing so would make it easier for the model to discern certain relationship of interest (subject-verb agreement) by shortening the path between relevant sub-units (subject and verb in a sentence)~\cite{kuncoro2018lstms,bjorne2009extracting}.
Hence, being explicit in modeling structures may be the key to obtaining performance gain.
Similarly, our work showed that modeling structures explicitly (using treeLSTM) is better than implicitly (using LSTM) in terms of model performance.

Models that learn task-specific trees from data could be better than models that use conventional parsers to obtain the trees~\cite{yogatama2017learning,choi2018learning}.
However, the learned trees are usually shallow and hard to interpret~\cite{williams2018latent}.
Shallow trees make the paths between related tokens shorter but they do not always result in better performance.
For the binary trees of logographs, the tree depth is unlikely to account for the improved performance because the trees are not balanced binary trees (which are shallowest).
The improved performance is more likely due to the inductive bias using logographic structures.
We showed that by exploiting structures like human intuition, treeLSTM could arrive at the general and correct solution in a more data-efficient and effective manner for pronunciation prediction and language modeling tasks concerning logographs (Chinese characters).
Better interpretability due to the model following human intuition provides some confidence that the model is general and is not exploiting statistical biases in the data.

\subsection{Potential Applications and Extensions}

To tackle the out-of-vocabulary problem, it is common to apply pre-processing steps such as replacing infrequent characters or characters unseen during training with the UNK token.
However, these pre-processing steps could potentially remove information stemmed from the usage of the infrequent characters.
Hierarchical embeddings enable modeling Chinese text directly without these pre-processing steps.
By treating Chinese characters as recursive structures of common sub-units instead of independent tokens, hierarchical embeddings make it possible for model to have a much bigger vocabulary.
Hierarchical embeddings also make learning representations of infrequent characters easier through leveraging the similarity between structures of infrequent characters and of common characters.
Thus, models that use hierarchical embeddings may be able to capture the intention behind the usage of infrequent characters.
Furthermore, hierarchical embeddings can also be used to model Japanese Kanji which are logographs created using the same principles as Chinese logographs.

We used hierarchical embeddings in the pronunciation prediction task and language modeling task.
However, other NLP tasks may also benefit from using hierarchical embeddings as previous work exploiting logograph structures have shown promising results in tasks such as machine translation~\cite{karpinska2018subcharacter,zhang2018neural,zhang2019chinese} or textual error detection~\cite{chen2015probabilistic}.
In particular, hierarchical embeddings may be useful in named entity recognition (NER) where infrequent characters may be used in names or in poetry generation where characters need to rhyme.

The current work can be extended in a few different ways. 
Although treeLSTM was used to construct the hierarchical embeddings, it is possible that self-attention models such as Transformer~\cite{vaswani2017attention} might lead to even better performance as they could learn patterns in trees that the current models could not.
However, since self-attention models usually have a lot of parameters they may overfit given the small amount of training data.
It would be interesting to see how well a bigger and more powerful model such as Transformer can model logographic structures given limited data.
For language modeling, using hierarchical embeddings does not constrain the choice of the language model.
The AWD-LSTM can be replaced by a more powerful model such as Transformer~\cite{vaswani2017attention} or BERT~\cite{devlin2019bert}.
It is possible that hierarchical embeddings may have little benefit for models such as Transformer or BERT which can exploit contextual information to deduce the representations of the infrequent characters.
However, exploiting both contextual information and structural similarity to obtain better embeddings for infrequent characters should theoretically be better than just relying on contextual information.

\section{Conclusion}

Exploiting recursive structures of logographs to build logographic embeddings can lead to embeddings that yield both better results and interpretability.
We showed both quantitative and qualitative evidence that exploiting recursive structures boosts accuracy in logographs' pronunciation prediction: the hierarchical embeddings is better than the embeddings constructed by LSTM, biLSTM, and CNN.
Hierarchical embeddings also consistently outperformed standard embeddings in language modeling of five different datasets.
Inspecting the inner workings of the models also revealed that treeLSTM conceivably resembles how humans perform reading tasks, suggesting that exploiting structures not only improves performance, but might also help us develop more interpretable models.
Although this paper only consider two tasks, building better logographic (character) embedding by exploiting recursive structures can potentially benefit other tasks such as machine translation.

\section*{Acknowledgment}
The authors would like to thank the anonymous reviewers for their constructive feedback to improve the paper.
In addition, the delightful discussions with Ai Ti Aw and Ed Hovy are also much appreciated.

\ifCLASSOPTIONcaptionsoff
  \newpage
\fi

\bibliographystyle{IEEEtran}
\bibliography{biblio,nlp}

\end{document}